%% file: samples/main.tex
\documentclass[sigconf]{acmart}

\usepackage[utf8]{inputenc}
\usepackage{threeparttable}  
\usepackage{booktabs}        
\usepackage{pifont}   
\usepackage{multirow}
\usepackage{float}
\usepackage{subfig}
\usepackage{graphicx}   
\usepackage{subcaption} 
\usepackage{hyperref}
\usepackage{pdfpages}
\usepackage{enumitem}
\usepackage{algorithm, algorithmicx, algpseudocode}

\makeatletter
\renewcommand{\paragraph}{%
  \@startsection{paragraph}{4}{0pt}
    {0.5ex \@plus 0.2ex \@minus 0.1ex}
    {-0.5em}
    {\normalfont\normalsize\bfseries}
}
\makeatother

\AtBeginDocument{%
  }

\setcopyright{acmlicensed}

\copyrightyear{2025}
\acmYear{2025}
\setcopyright{acmlicensed}\acmConference[MM '25]{Proceedings of the 33rd ACM International Conference on Multimedia}{October 27--31, 2025}{Dublin, Ireland}
\acmBooktitle{Proceedings of the 33rd ACM International Conference on Multimedia (MM '25), October 27--31, 2025, Dublin, Ireland}
\acmDOI{10.1145/3746027.3754985}
\acmISBN{979-8-4007-2035-2/2025/10}

\begin{document}
\bibliographystyle{plain}

\title{SeqVLM: Proposal-Guided Multi-View Sequences Reasoning via VLM for Zero-Shot 3D Visual Grounding}


\author{Jiawen Lin\textsuperscript{*}}
\affiliation{%
    \institution{School of Informatics, Xiamen University}
    \city{Xiamen}
    \state{Fujian}
    \country{China}
}
\email{linjiawen@stu.xmu.edu.cn}

\author{Shiran Bian\textsuperscript{*}}
\affiliation{%
    \institution{School of Informatics, Xiamen University}
    \city{Xiamen}
    \state{Fujian}
    \country{China}
}
\email{bianshiran@foxmail.com}

\author{Yihang Zhu}
\affiliation{%
    \institution{School of Computer Science, Nanjing University}
    \city{Nanjing}
    \state{Jiangsu}
    \country{China}
}
\email{zhu_yih@163.com}

\author{Wenbin Tan}
\affiliation{%
    \institution{School of Informatics, Xiamen University}
    \city{Xiamen}
    \state{Fujian}
    \country{China}
}
\email{wbtan@stu.xmu.edu.cn}

\author{Yachao Zhang\textsuperscript{\dag}}
\affiliation{%
    \institution{School of Informatics, Xiamen University}
    \city{Xiamen}
    \state{Fujian}
    \country{China}
}
\email{yachaozhang@xmu.edu.cn}

\author{Yuan Xie}
\affiliation{%
    \institution{School of Computer Science and Technology, East China Normal University}
    \city{Shanghai}
    \country{China}
}
\email{yxie@cs.ecnu.edu.cn}

\author{Yanyun Qu\textsuperscript{\dag}}
\affiliation{%
    \institution{Key Laboratory of Multimedia Trusted Perception and Efficient Computing, Ministry of Education of China, Xiamen University}
    \city{Xiamen}
    \state{Fujian}
    \country{China}
}
\email{yyqu@xmu.edu.cn}

\renewcommand{\shortauthors}{Jiawen Lin et al.}

\begin{abstract}
3D Visual Grounding (3DVG) aims to localize objects in 3D scenes using natural language descriptions. Although supervised methods achieve higher accuracy in constrained settings, zero-shot 3DVG holds greater promise for real-world applications since eliminating scene-specific training requirements. However, existing zero-shot methods face challenges of spatial-limited reasoning due to reliance on single-view localization, and contextual omissions or detail degradation. To address these issues, we propose SeqVLM, a novel zero-shot 3DVG framework that leverages multi-view real-world scene images with spatial information for target object reasoning. Specifically, SeqVLM first generates 3D instance proposals via a 3D semantic segmentation network and refines them through semantic filtering, retaining only semantic-relevant candidates. A proposal-guided multi-view projection strategy then projects these candidate proposals onto real scene image sequences, preserving spatial relationships and contextual details in the conversion process of 3D point cloud to images. Furthermore, to mitigate VLM computational overload, we implement a dynamic scheduling mechanism that iteratively processes sequances-query prompts, leveraging VLM's cross-modal reasoning capabilities to identify textually specified objects. Experiments on the ScanRefer and Nr3D benchmarks demonstrate state-of-the-art performance, achieving Acc@0.25 scores of 55.6\% and 53.2\%, surpassing previous zero-shot methods by 4.0\% and 5.2\%, respectively, which advance 3DVG toward greater generalization and real-world applicability. The code is available at \url{https://github.com/JiawLin/SeqVLM}.
\end{abstract}






\maketitle

\begingroup
\renewcommand\thefootnote{*}
\footnotetext{Jiawen Lin and Shiran Bian contributed equally to this work.}
\renewcommand\thefootnote{\dag}
\footnotetext{Yanyun Qu and Yachao Zhang are corresponding authors.}
\endgroup

\section{Introduction}

\begin{figure}[htbp]
    \centering
    \vspace{1em}
    \includegraphics[width=1\linewidth]{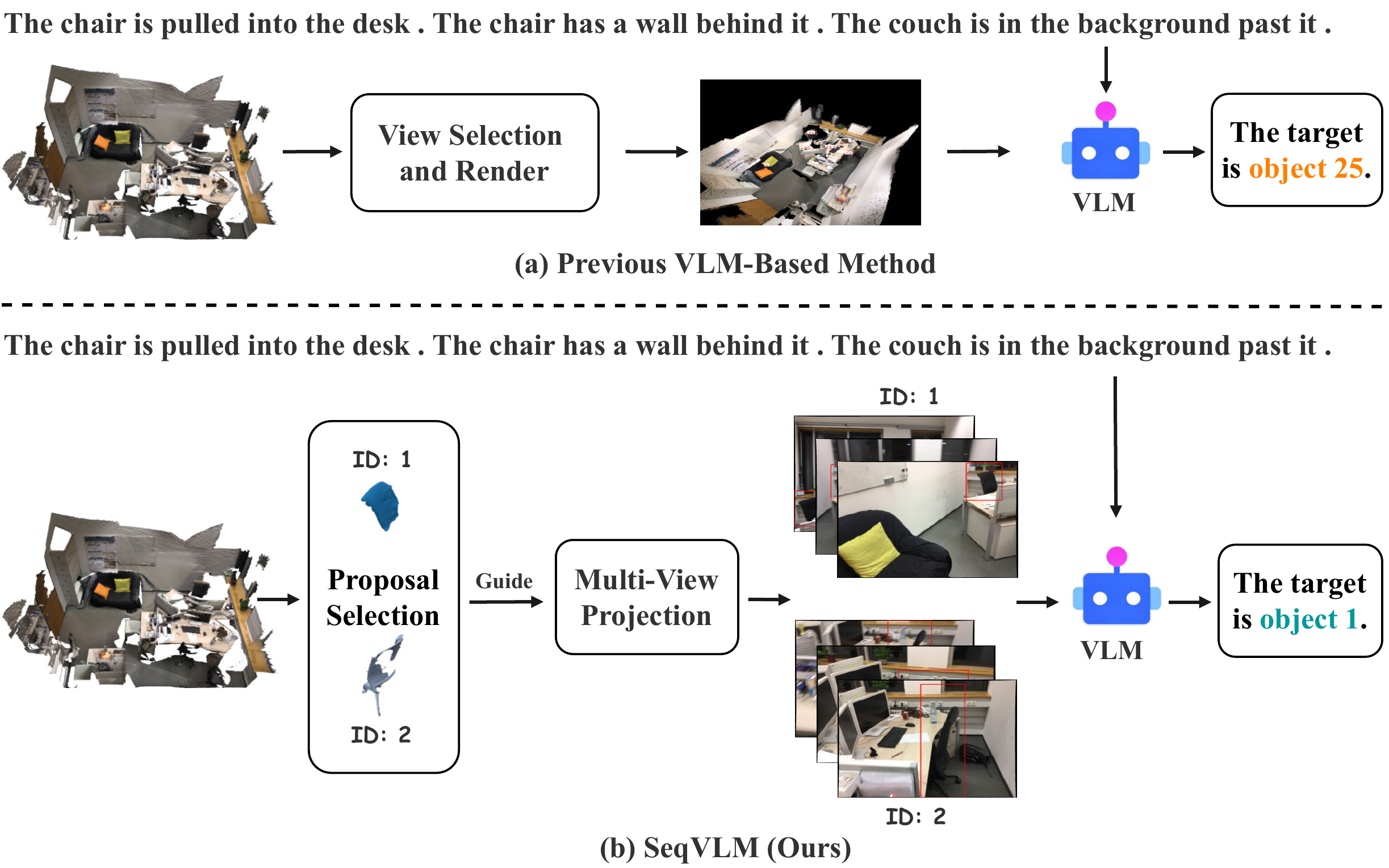}
    \caption{Comparative overview of previous VLM-based methods and SeqVLM. (a) Previous VLM-based methods for 3DVG use single-view images for localization. (b) SeqVLM integrates proposal-guided with multi-view real-image projection for accurate 3DVG.}
    \label{fig:motivation}
\end{figure}

3D Visual Grounding (3DVG) is a critical task that aims to establish cross-modal alignment between natural language descriptions and target objects in 3D scenes. Given a 3D scene represented as point clouds and a semantically rich textual query(\textit{e.g., "the red chair near the window"}), the task requires a model to precisely infer the 3D bounding box coordinates of the referred object. This technology holds significant application value in intelligent human-robot interaction \cite{li2024seeground,liu2024aligning}, autonomous driving environmental perception \cite{zhou2022context, 3dvg-trans}, and AR/VR systems\cite{zhang2024towards,unal2024four}.

Current approaches predominantly rely on fully supervised paradigms \cite{scanrefer, tgnn, sat,mcln,eda,3dsps,mvt,3dvlp}, achieving localization through training on fixed text-scene paired datasets. Although these methods attain high accuracy in closed datasets, they suffer from two limitations: (1) High annotation costs for 3D bounding boxes restrict training data scale, hindering generalization to real-world scene complexity \cite{huang2024training,hwang2023joint,lin2024exploring}; (2) Strict dependence on annotated data impedes adaptation to open-vocabulary scenarios \cite{smith2023flowcam,deitke2023objaverse}. 

Zero-shot and open-vocabulary learning provide promising directions to address these limitations \cite{llm-grounder,vlmgrounder,zs3dvg,li2024seeground,openscene}. Zero-shot learning enables models to localize unseen objects without training, while open-vocabulary learning focuses on constructing models capable of processing arbitrary vocabulary. The integration of these paradigms enhances model generalization and broadens application scenarios. Concurrently, capitalizing on the profound semantic and scene-level understanding of large language models (LLMs) \cite{zhang2024earthgpt,wu2024visionllm,chen2023minigpt,wang2023chat,huang2025reason3d}and visual-language models (VLMs)\cite{zheng2023preventing,javed2024cplip,phan2024zeetad,chen2024reasoning3d}, researchers are developing zero-shot 3D visual grounding frameworks to tackle open-vocabulary challenges.

Existing zero-shot 3DVG methods primarily fall into two categories: LLM-based and VLM-based methods. LLM-based methods \cite{llm-grounder,zs3dvg} convert textual descriptions into structured programmatic instructions via LLMs, then execute code-based object retrieval in point cloud space. While these methods leveraging LLMs' superior semantic comprehension to parse complex language structures, they struggle with sparse point clouds' inherent limitations in capturing color/texture cues and reasoning about intricate spatial relationships. Conversely, as shown in Figure \ref{fig:motivation}(a), previous VLM-based methods \cite{vlmgrounder,li2024seeground} enhance 3D visual grounding by integrating 2D visual features to identify objects in images that best match text descriptions. However, their reliance on single-view renderings may introduce cascading limitations. The absence of 3D geometric constraints leads to spatial biases that misalign 2D projections with 3D coordinates, while single-view perspectives inherently fail to resolve occlusions or capture multi-object contextual relationships. These issues are further compounded by gaps between rendered and real-world imagery, such as chromatic aberrations, texture simplification, and occlusion \cite{yan2024multi,cao2024tparn}, which systematically degrade the VLMs’ ability to reason about fine-grained visual-textual correspondences \cite{li2023blip,liu2024unimel,zhang2024prototype}. Collectively, these limitations highlight the critical need for geometrically consistent, multi-view fusion strategies in real-world 3D grounding scenarios. 


To address these challenges, we propose SeqVLM, a novel framework that enhances cross-modal alignment for zero-shot 3DVG by leveraging multi-view real-world image sequences. SeqVLM integrates 3D point clouds, multi-view images, and natural language descriptions, utilizing VLM for cross-modal alignment to achieve precise 3D object localization. The overall pipeline of SeqVLM is illustrated in Figure \ref{fig:motivation}(b).

Specifically, our framework uses a 3D semantic segmentation network to extract object proposals from 3D point cloud scenes. To improve accuracy and efficiency, irrelevant proposals are filtered through semantic alignment with the target object category described in the input, retaining only those that match semantically with the specified category. The refined proposals are then projected onto real-world image sequences to generate context-aware regions with multi-view consistency, preserving spatial relationships and details. To address the constraints imposed by VLM input length, we introduce an iterative reasoning mechanism that dynamically schedules inference, optimizing both inference efficiency and localization accuracy. Compared to existing approaches, SeqVLM has two main advantages: First, proposal-guided projection strategy effectively maintains 3D geometric attributes and environmental context, significantly reducing localization errors. Second, by leveraging multi-view real images, SeqVLM enhances spatial understanding and contextual awareness, thereby enhancing the cross-modal alignment of VLMs. These innovations enable robust zero-shot 3DVG in complex environments.


SeqVLM achieves absolute improvements of 4.0\% and 5.2\% in Acc@0.25 on ScanRefer \cite{scanrefer} and Nr3D \cite{referit3d}, respectively, outperforming existing zero-shot methods. Notably, its performance rivals fully supervised approaches on certain metrics. 

Overall, our contributions are summarized as follows:
\begin{itemize}
    \item[$\bullet$] We propose SeqVLM, a novel framework that integrates 3D geometric features with 2D visual cues via proposal-guided multi-view projection, enabling accurate cross-modal localization in zero-shot settings.
    \item[$\bullet$]  We develop a proposal-guided multi-view projection strategy, which aligns objects semantically relevant to textual descriptions with real-scene images across multiple viewpoints. This ensures precise projection positioning while preserving contextual and detailed scene information.
    \item[$\bullet$] We design an iterative reasoning mechanism to address VLM failure in multi-candidate scenarios through dynamic computational scheduling.
    \item[$\bullet$] We conduct extensive experiments on the ScanRefer and Nr3D datasets, demonstrating the effectiveness of our framework and setting new state-of-the-art performance in zero-shot 3D visual grounding.
\end{itemize}


\begin{figure*}[htbp]
    \centering
    \includegraphics[width=\linewidth]{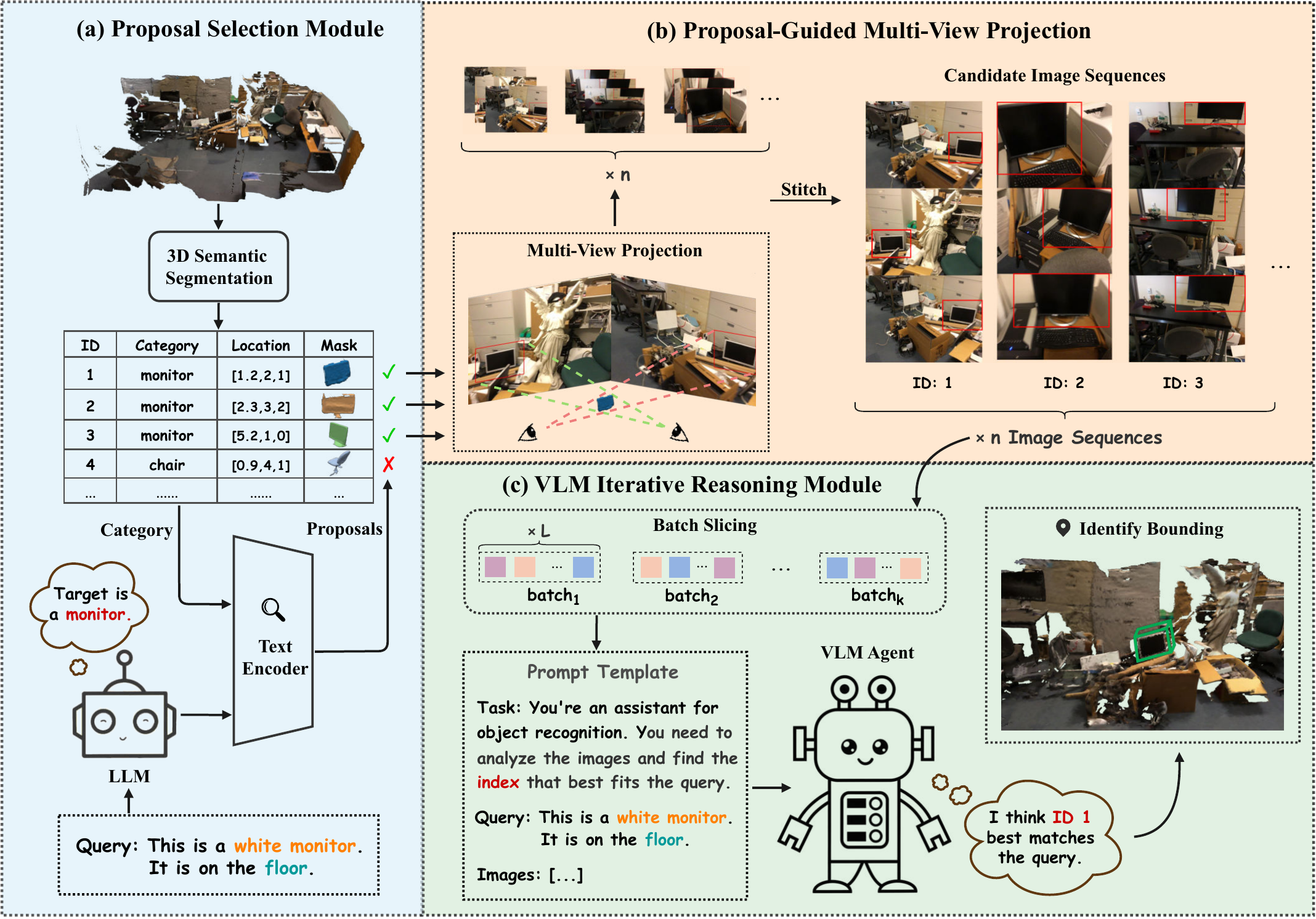}
    \caption{Overview of SeqVLM. Given a 3D scene and a textual query, SeqVLM localizes the target object through a structured pipeline comprising three key modules. First, the Proposal Selection Module employs a 3D semantic segmentation network combined with semantic filtering to extract candidate proposals aligned with the query. Next, the Proposal-Guided Multi-View Projection Module projects these candidates onto optimally selected 2D views, incorporating multi-view stitching to preserve spatial and contextual details. Finally, the VLM Iterative Reasoning Module iteratively refines the search space via a dynamic reasoning mechanism to identify the most probable target. The final selection is then mapped back to its bounding box, ensuring robust localization in complex environments.}
    \label{fig:framework}
\end{figure*}

\section{Related Work}

\subsection{Supervised 3DVG}
Current research in 3D visual grounding predominantly adopts supervised learning paradigms, achieving cross-modal alignment through training on annotated datasets. Existing methodologies can be categorized into two technical frameworks: two-stage and single-stage approaches.

Two-stage methods \cite{scanrefer,referit3d,3dvg-trans,jain2022bottom,chen2022d} employ a "detect-then-match" cascade architecture. The first stage generates candidate object proposals via 3D object detection or instance segmentation\cite{liu2021group,jiang2020pointgroup,vu2022softgroup}, while the second stage constructs cross-modal alignment modules to refine language-query matching with candidate region features and optimize bounding boxes\cite{3djcg,3dsps}. In contrast, single-stage methods \cite{eda,wang2024g, mcln} utilize end-to-end architectures to jointly model 3D point clouds and textual features, achieving dense regression predictions of target locations. Notably, SAT \cite{sat} pioneers the integration of 2D image semantics as auxiliary supervision during training, enhancing joint representation learning between 3D point clouds and language via cross-modal alignment, thereby significantly improving localization accuracy. Despite remarkable progress, supervised approaches face two critical bottlenecks: (1) High annotation costs for 3D data restrict dataset scalability, limiting coverage of complex real-world scenarios; (2) Model generalization is constrained by closed-vocabulary settings, hindering adaptation to open-world localization of unseen objects.

\subsection{Zero-Shot 3DVG}
Zero-shot methods aim to overcome the annotation dependency of supervised paradigms by leveraging the cross-modal reasoning capabilities of pretrained large models, primarily divided into Large Language Model (LLM)-based and Visual-Language Model (VLM)-based methods.

LLM-based methods \cite{llm-grounder,zs3dvg} parse natural language queries through linguistic models, converting them into structured semantic elements or executable program instructions. These methods combine geometric reasoning with commonsense knowledge for zero-shot localization. While effective in handling complex semantic inference, they exhibit limitations in exploiting 3D scene contextual information and effectively integrating fine-grained visual features.

VLM-based methods \cite{vlmgrounder,li2024seeground}, leverage the multimodal alignment capabilities of visual-language models, enabling the integration of visual and textual information for enhanced scene understanding. VLM-Grounder \cite{vlmgrounder} proposes dynamically stitching multi-view 2D image sequences, utilizing VLM's open-vocabulary reasoning to infer 2D object masks, followed by multi-view projection and morphological processing for 3D bounding box reconstruction. SeeGround \cite{li2024seeground} further introduces dynamically rendered 2D images with spatially enriched 3D text descriptions, utilizing a Perspective Adaptation Module for query-aligned views and a Fusion Alignment Module to improve object localization. Despite improved scene context utilization, these methods remain constrained by some key factors: (1) Geometric misalignment errors arising from improper projection methodologies; (2) Loss of spatial information related to viewpoint-specific cues due to single-view constraints; (3) Visual feature distortion due to domain gaps between rendered images and real-world scenes.

\section{Methodology}
\subsection{Task Definition}
We consider a 3D scene represented by a colored point cloud $\mathbf{P} \in \mathbb{R}^{N \times 6}$, where each point is characterized by XYZ coordinates and RGB color attributes, along with an image list captured from various viewpoints in real-world scenes, and a textual description $T$ specifying the target object $O^{*}$. Our task focuses on zero-shot 3DVG that localizes the $O^{*}$ in 3D space without requiring scene-specific training or fine-tuning. The image list can be obtained through various sensors including RGB-D sensors with structured light scanners \cite{shen2013layer, micko2025motion}, dense SLAM systems \cite{hu2024cg, huang2024photo, li2024rgb} or Time-of-Flight (ToF) sensors \cite{xu2024m,qiao2024rgb}.

\subsection{Overview}
The framework of SeqVLM is illustrated in Figure \ref{fig:framework}. Given a point cloud scene $\mathbf{P}$, a 3D semantic segmentation network is first employed to perform instance segmentation, followed by confidence-based filtering to extract object categories and corresponding mask features. Then, LLM parses the textual description $T$ to identify the target object category, denoted as $C^*$. This category, along with the filtered object categories, is embedded using a text encoder. Based on the cosine similarity matching, a proposal list $\mathcal{O}$ is generated, containing $n$ proposals that are semantically related to the target category. For each proposal $O_i \in \mathcal{O}$, all scene images from $\mathbf{P}$ are sampled to form an image list $\mathcal{I}_i$. Multi-view projection is applied to annotate the region corresponding to $O_i$ in each image within $\mathcal{I}_i$. The top $n_{\mathit{frame}}$ images with the largest projected areas are then stitched, generating a vertical image sequence ${S}_i$ as the 2D visual representation of $O_i$. Finally, using an iterative reasoning mechanism, the proposal sequences $\mathcal{S} = [{S}_1, {S}_2, \dots, {S}_k]$, together with the textual description $T$, are fed into the VLM for cross-modal reasoning to identify the target object $O^*$. By deeply integrating multimodal information, SeqVLM significantly enhances localization accuracy and robustness, offering an effective solution for zero-shot object localization in complex 3D scenes.

\subsection{Proposal Selection Module}
We propose to employ a text-driven semantic filtering mechanism to refine proposals whose categories are consistent with the target object. Specifically, the module processes the point cloud scene $\mathbf{P}$ using a 3D semantic segmentation network $\Phi$, and retain only those instances with confidence scores exceeding a predefined threshold $\theta$:
\begin{equation}
    \Phi(\mathbf{P}) = \{M_i \mid \sigma(M_i) \ge \theta \}_{i = 1}^{m},
\end{equation}
where $M_i$ and $\sigma(\cdot)$ denote the instance mask features and confidence function, respectively. The adoption of threshold filtering effectively eliminates the unreliable segmentation result.

Following the dynamic vocabulary mapping in ZSVG3D \cite{zs3dvg}, a structured Object Profile Table($\mathcal{OPT}$) can be constructed as:
\begin{equation}
    \mathcal{OPT} = \{ (\textit{ID}_i, C_i, BBox_i, M_i) \}_{i = 1}^m, 
\end{equation}
where $C_i$ represents the object category and $BBox_i$ corresponds to the 3D bounding box. Then the text encoder is applied to compute the embedding of proposal category $C_i$ and the $T_{\mathit{target}}$ generated by LLM:
\begin{equation}
    \begin{aligned}[c]
        & E_t = f_{\mathit{text}} (T_{\mathit{target}}), \\
        & E_c^{(i)} = f_{\mathit{text}}(C_i),
    \end{aligned}
\end{equation}
where $f_{\mathit{text}}(\cdot)$ is the text encoder, and the cosine similarity is computed as:
\begin{equation}
    \zeta_i = \frac{E_t \cdot E_c^{(i)}}{\left \| E_t \right \| \cdot \left \| E_c^{(i)} \right \|}. 
\end{equation}

The target object category is determined by the maximum similarity, which then defines the proposal set $\mathcal{O}$ as:

\begin{equation}
    C^{*} = \mathop{\arg\max} \limits_{C_i \in \mathcal{OPT}} \zeta_{i},
\end{equation}
\begin{equation}
    \mathcal{O} = \{ M_i \mid C_i = C^{*} \}_{i = 1}^{n}.
\end{equation}
    
This optimization retains only proposals that share the same category as the target, significantly reducing computational complexity in VLM reasoning.

\subsection{Proposal-Guided Multi-View Projection}
Since VLM cannot process 3D scenes properly, we propose a multi-view projection method that transforms 3D point clouds into 2D images to adapt to VLM. However, in previous works, VLM-Grounder \cite{vlmgrounder} directly maps 2D detection results to 3D space without considering geometric constraints, and SeeGround \cite{li2024seeground} relies on restricted single-view rendering, resulting in inadequate spatial understanding and detail loss. Both approaches lack the full utilization of proposal-guided images. To address these limitations, our multi-view projection method is guided by proposals to preserve the context of objects.

\paragraph{Proposal-Guided Projection.} For one point belonging to a proposal, its world coordinate can be defined as $P_w = [x_w, y_w, z_w]$, which can be projected into the camera coordinate system of the corresponding view via the homogeneous transformation matrix $T_{\mathit{wc}} \in \mathbb{R}^{4 \times 4}$:
\begin{align}
    P_c &= T_{\mathit{wc}} \cdot  [x_w, y_w, z_w, 1]^{\mathrm{T}}
        = [x_c, y_c, z_c, 1]^{\mathrm{T}}.
\end{align}

The pixel coordinates $(u, v)$ in the view image are then computed using the intrinsic matrix
$\mathbf{K} = 
\begin{bmatrix}
    f_x & 0 & c_x \\
    0 & f_y & c_y \\
    0 & 0 & 1
\end{bmatrix}$:
\begin{equation}
    u = \frac{x_c \cdot f_x}{z_c} + c_x,\ v = \frac{y_c \cdot f_y}{z_c} + c_y.
\end{equation}

To ensure projection validity, we introduce a depth consistency verification:
\begin{equation}
    \left | \frac{D(u, v) - z_c}{D(u, v)} \right | \le \tau ,
\end{equation}
where $D(u,v)$ denotes the depth measurement at $(u,v)$ in the view image, and $\tau$ serves as the visibility threshold. If $P_w$ satisfies the consistency after transformation, it is considered valid and visible in the corresponding view.

\paragraph{Multi-View Sequence Generation.}For each proposal $M_i \in \mathcal{O}$, which contains a set of point cloud coordinates, we optimize viewpoint selection by quantifying the projected area. The top $n_{\mathit{frame}}$ images with the largest projected areas are selected from the sampled image set $\mathcal{I}_i$, forming the pixel projection set $ \mathcal{P}_i = \{ p_i^{1}, p_i^{2}, \dots, p_i^{n_{\mathit{frame}}} \}$, where each element represents the pixel projection set of $M_i$ from a certain viewpoint.

During the image stitching phase, each pixel set $p_i^{j} \in \mathcal{P}_i$ is analyzed to extract its minimum bounding rectangle in the view image:
\begin{equation}
    \mathcal{R} = [u_{\min}, v_{\min}, u_{\max}, v_{\max}],\ (u, v) \in p_{i}^{j},
\end{equation}
which represents the 2D bounding box of the proposal. To preserve the image details of the object within the bounding box, the box is expanded by the same proportion $\alpha$ along both the width and height directions, and is then highlighted by a distinct 3-pixel-wide red rectangular annotation.

After that, $n_{\mathit{frame}}$ annotated images are vertically concatenated to generate an enhanced image sequence $S_i$. This process avoids the location deviation during grounding, and enriches spatial relationships and contextual details of object through multiple viewpoints, which are essential for robust 3D scene understanding by VLM.

\subsection{VLM Iterative Reasoning Module}
We propose an iterative reasoning mechanism to address the sensitivity of VLM to input length in zero-shot 3DVG tasks. Existing works \cite{vlmgrounder, zhao2024stitch,wang2024survey} indicate that there is a trade-off between computational load and reasoning accuracy when VLM analyzes a large number of high-resolution images. If we input all image sequences of $\mathcal{S}$ in a single round of dialogue, it may overload the VLM, degrade performance, or even cause response timeouts. Therefore, we slice the image sequences $\mathcal{S}$ into batches, allowing VLM to optimize the set of proposals over rounds, thus achieving precise localization of the target object.

\input{samples/algos/vlm_predict}

The details of the mechanism are illustrated in Algorithm \ref{algo:vlm-predict}. In this algorithm, the variable $\mathcal{Q}$ is defined to dynamically maintain the image sequences that are to be analyzed. Before the iteration process begins, $\mathcal{Q}$ will be sliced into batches $\mathcal{B} = \{ B_1, B_2, \dots, B_q \}$ with a maximum batch size $L$. Subsequently, each batch is combined with the textual description $T$ to form the input prompt for the VLM. If the VLM identifies a unique image sequence $S^* \in B_i$ that best matches $T$, $S^*$ will be reinserted to $\mathcal{Q}$; otherwise, $B_i$ will be discarded.

After each reasoning step, the algorithm checks the state of $\mathcal{Q}$. The iteration will stop only when $\lvert \mathcal{Q}\rvert$ is less than or equal to one. Finally, if only one image sequence remains in $\mathcal{Q}$, its corresponding index is selected as the final result. This mechanism circumvents the limitations of VLM in long-sequence reasoning by gradually reducing the search space. 

For 3D localization, the bounding box of selected proposal can be retrieved from the Object Profile Table ($\mathcal{OPT}$), completing the end-to-end 3DVG pipeline.

\input{samples/tables/scanrefer}
\input{samples/tables/nr3d}

\section{Experiments}
\subsection{Experimental Settings}

\paragraph{Datasets.}
We evaluate our framework on two established 3D visual grounding benchmarks: ScanRefer\cite{scanrefer} and Nr3D\cite{referit3d}. ScanRefer, constructed from ScanNet, comprises 51,583 natural language descriptions paired with 11,046 objects across 800 indoor scenes, emphasizing fine-grained alignment between linguistic expressions and 3D object localization through attribute and spatial-relation reasoning. The dataset is divided into Unique and Multiple subsets: Unique targets objects uniquely identifiable by intrinsic attributes, while Multiple requires disambiguation among multiple instances of the same category using spatial relations. Nr3D contains 41,503 descriptions for 7,189 objects across 1,448 scenes, prioritizing viewpoint-dependent language diversity in real-world contexts. Its evaluation categorizes references into Easy and Hard subsets based on the necessity of implicit contextual reasoning, as well as View-Dependent and View-Independent subsets to assess robustness to observer perspectives. 

\begin{figure*}
    \centering
    \includegraphics[width=1\linewidth]{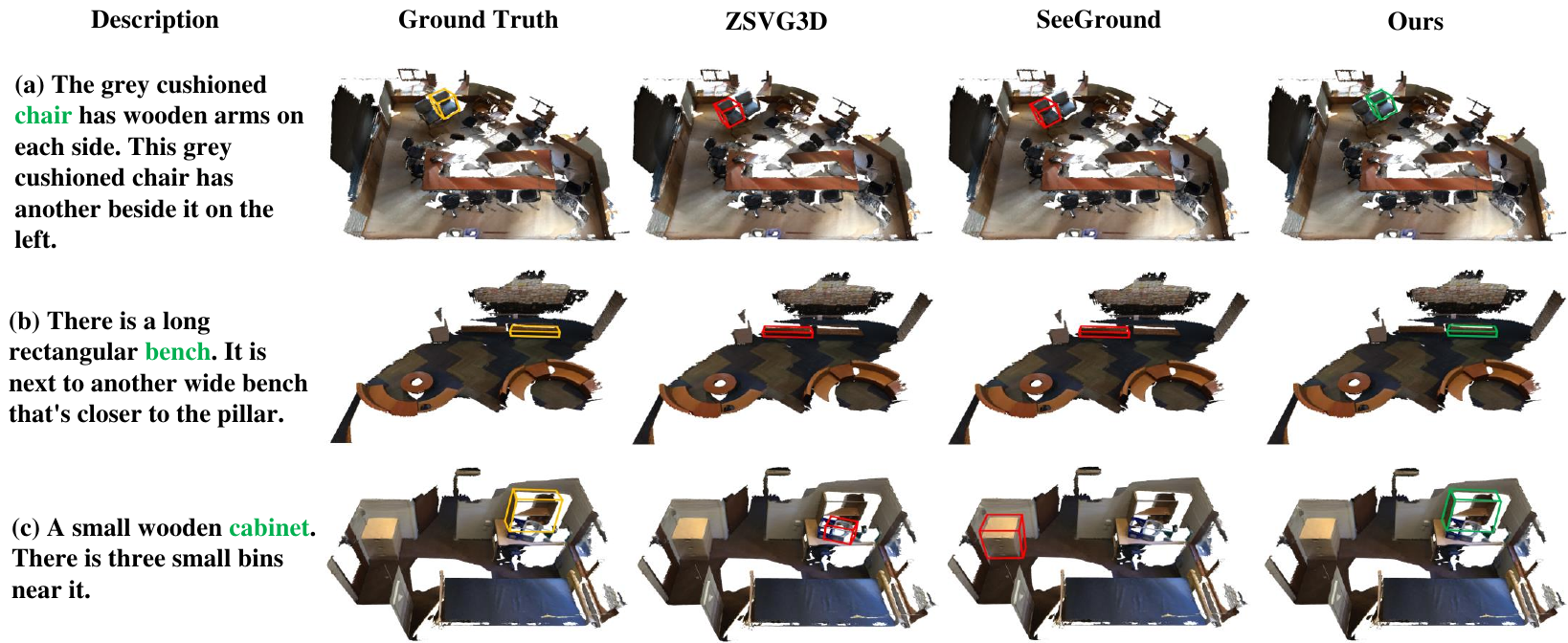}
    \caption{ Qualitative results of 3D visual grounding on ScanRefer\cite{scanrefer} dataset. Rendered images of 3D scans are presented, including the ground-truth (yellow),
 incorrectly identified objects (red), and correctly identified objects (green).}
    \label{fig:qualitative}
\end{figure*}

\paragraph{Implementation Details.}
Experiments were conducted on a TITAN RTX GPU using Doubao-1.5-pro\cite{volcengine2025} as the LLM for semantic-guided proposal selection and Doubao-1.5-vision-pro\cite{volcengine2025} as the VLM for multimodal reasoning. We employ Mask3D \cite{mask3d} as the 3D semantic segmentation network, applying a confidence threshold of 0.2 to filter low-quality segmentation masks. Semantic matching is performed using the CLIP-ViT-Base-Patch16 text encoder \cite{radford2021learning}. Scene image sequences are segmented into 20-frame intervals for viewpoint sampling, with each candidate object projected across five optimized viewpoints validated through depth consistency checks (visibility threshold is 0.25). We set the bounding region expansion parameter $\alpha$ to 0.25. The VLM processes candidates iteratively through an iterative reasoning mechanism with a batch size limit of four. Due to the high computational cost of VLM-based models, we follow the standardized protocol of VLM-Grounder \cite{vlmgrounder} for fair evaluation and reproducibility, testing on 250 validation samples per benchmark dataset, matching previous evaluation conditions.

\subsection{Quantitative Results}

\paragraph{ScanRefer.}

SeqVLM establishes a new state-of-the-art in zero-shot 3D visual grounding, as demonstrated by the comprehensive comparisons in Table~\ref{tab:scanrefer_result}, while maintaining competitive performance relative to fully-supervised methods. Specifically, SeqVLM achieves 55.6\% Acc@0.25 and 49.6\% Acc@0.5 on the overall benchmark, outperforming previous state-of-the-art zero-shot methods, VLM-Grounder\cite{vlmgrounder} and SeeGround\cite{li2024seeground}, by +4.0\% and +10.2\%, respectively. In single-object scenarios (Unique subset), SeqVLM surpasses prior approaches with gains of 1.6\% and 3.8\% in Acc@0.25 and Acc@0.5, respectively, highlighting its superior localization accuracy. For multi-object scenarios (Multiple subset), while Acc@0.25 marginally lags behind VLM-Grounder, our method achieves the highest Acc@0.5 performance (+7.8\%), demonstrating enhanced fine-grained localization capabilities. Notably, the framework exhibits competitive results against fully-supervised baselines while maintaining strong generalization across diverse 3D visual grounding tasks. These results validate the effectiveness of the proposed method and underscore its potential for practical applications requiring robust 3D spatial reasoning.

\begin{figure}
    \centering
    \includegraphics[width=1\linewidth]{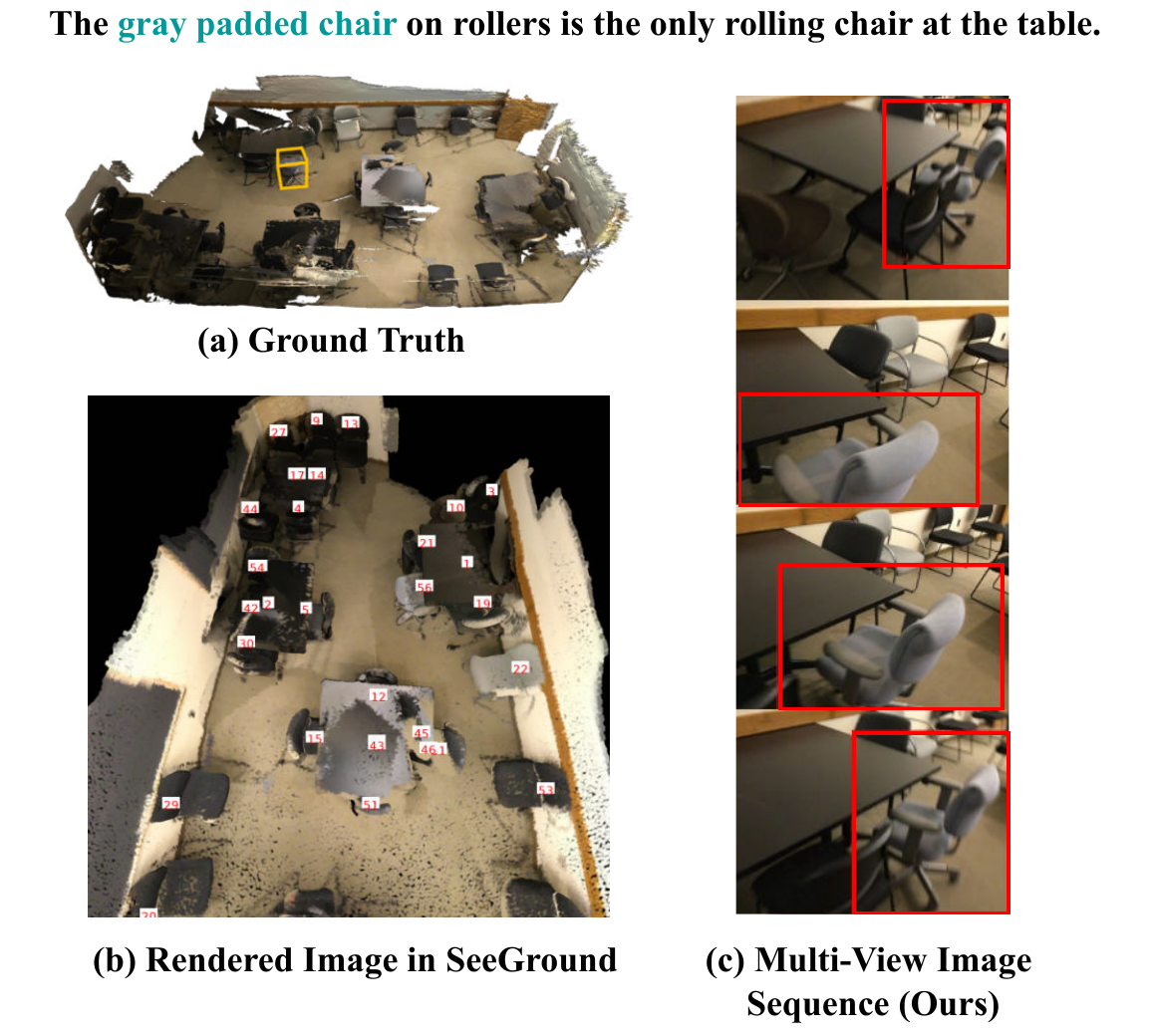}
    \caption{Comparative visualization of single-view rendering and multi-view sequence.}
    \label{fig:vs_seeground}
\end{figure}

\paragraph{Nr3D.}
As shown in Table ~\ref{tab:nr3d_result}, SeqVLM attains an overall accuracy of 53.2\% on the Nr3D dataset, exceeding prior state-of-the-art zero-shot methods by 5.2\%. Our method exhibits robust performance across varying scenario complexities, achieving 58.1\% (+2.9\%) and 47.4\% (+7.9\%) accuracy in the "Easy" and "Hard" subsets, respectively. Equally notable is its performance in view-dependent and view-independent scenarios, with accuracies of 51.0\% (+5.2\%) and 54.5\% (+5.1\%), illustrating its effective handling of perspective-dependent variations.

\subsection{Qualitative Results}
Figure \ref{fig:qualitative} presents the qualitative 3D visual grounding results of ZSVG3D\cite{zs3dvg}, SeeGround \cite{li2024seeground}, and SeqVLM on the ScanRefer \cite{scanrefer} dataset. Cases (a–c) demonstrate the superior localization capability of our model, even in complex environments with challenging textual descriptions. For instance, in case (a), our model accurately identifies the target chair despite multiple chairs being present in the scene. In contrast, both baseline methods exhibit lower localization accuracy, with ZSVG3D occasionally misclassifying object categories (e.g., identifying bins instead of cabinets in case (c)). These qualitative results underscore the efficacy of our approach in enhancing grounding precision and robustness.

Figure \ref{fig:vs_seeground} contrasts the visual inputs provided to VLM by See-Ground \cite{li2024seeground} and our method. The ground-truth (a) depicts \textit{“the gray padded chair on rollers. it is the only rolling chair at the table.”} However, SeeGround’s single-view rendering (b) occludes these distinguishing features due to viewpoint constraints, resulting in misidentification. In contrast, our multi-view sequence (c) integrates complementary perspectives, resolving occlusions and clearly revealing the wheels. This visual evidence directly accounts for the accurate localization of our method, demonstrating that multi-view fusion mitigates single-view ambiguities inherent in real-world 3D grounding tasks.

\subsection{Ablation Studies}

\paragraph{Component Ablation.}
To evaluate the contribution of core components in our framework, comprehensive ablation studies are conducted on the ScanRefer benchmark as shown in Table ~\ref{tab:ablation_module}. The baseline model achieves merely 2.4\% accuracy under the Acc@0.5 metric, highlighting the inherent complexity of zero-shot 3D visual grounding. Introducing the Proposal Selection Module significantly improves performance to 43.2\% by filtering semantically relevant candidates through text-driven category alignment, whereas standalone implementations of the Multi-View Projection technique or Iterative Reasoning mechanism yield limited gains of 2.0\% and 10.0\% respectively. The complete framework integrating all components attains 49.6\% accuracy, demonstrating substantial synergistic effects through the combination of semantic-aware multi-view 2D visual representations and iterative candidate refinement. This progression underscores the indispensability of each module, particularly the Proposal Selection Module which contributes the most critical performance leap by bridging 3D segmentation with language semantics.

\input{samples/tables/abla_module}
\input{samples/tables/abla_vlm}

\input{samples/tables/abla_cross_method}

\begin{figure}
    \centering
    \includegraphics[width=1\linewidth]{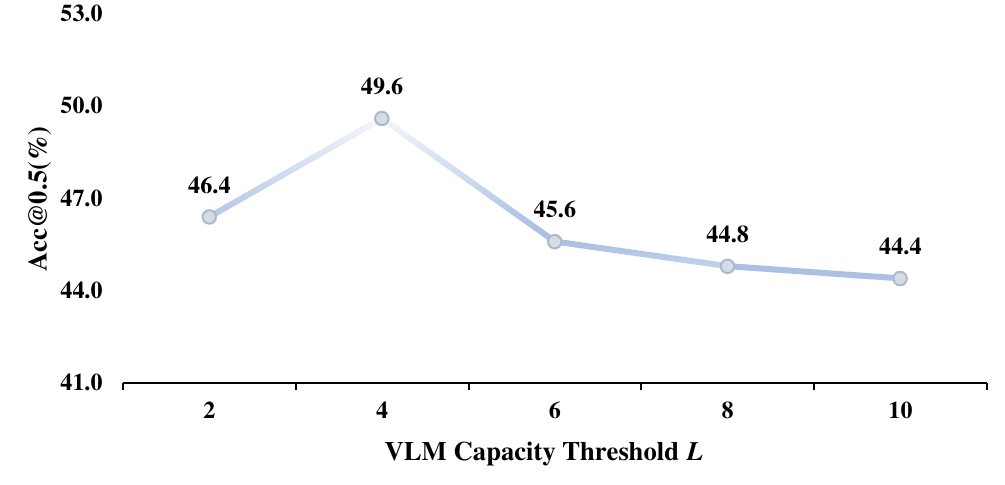}
    \caption{Ablation study on different values of batch size threshold $L$.}
    \label{fig:ablation_L}
\end{figure}

\begin{figure}
    \centering
    \includegraphics[width=1\linewidth]{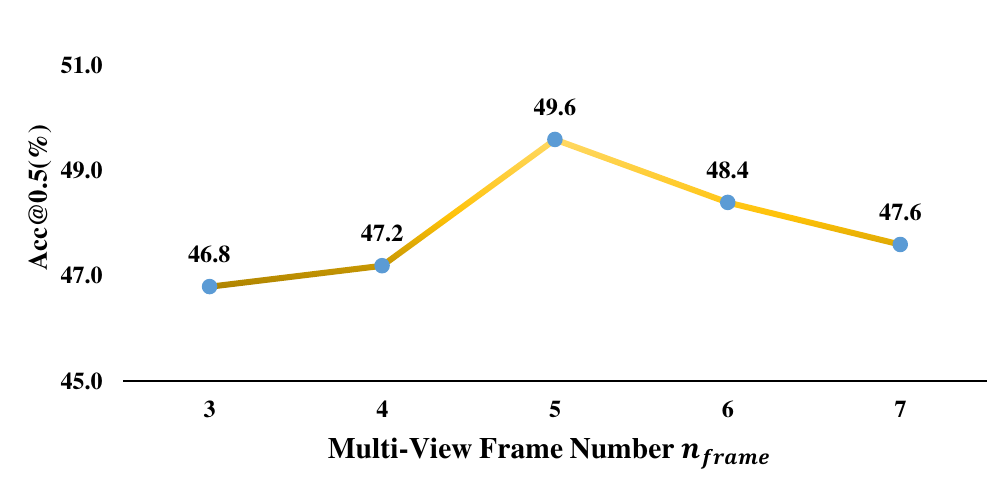}
    \caption{Ablation study on different values of multi-view frame number $n_{\mathit{frame}}$.}
    \label{fig:ablation_nframe}
\end{figure}

\paragraph{VLM Selection.}
We evaluate the influence of different VLMs on grounding performance and cost. As shown in Table~\ref{tab:ablation_vlm}, GPT-4\cite{chatgpt2024} achieves 44.0 Acc@0.5 with moderate token usage (1,684k) and cost (2.2\$), while Qwen-vl-max\cite{qwen2025} improves accuracy to 46.0 at a lower cost (0.8\$). Doubao-1.5-vision-pro achieves the highest accuracy (49.6 Acc@0.5) with higher token usage (7,097k), indicating superior visual-text alignment. Given the importance of precision in 3DVG, we choose Doubao as the default VLM, favoring its accuracy despite increased computational cost.

\paragraph{Cross-Method Comparison.} 
To decouple the impact of framework design from VLM selection, we conduct cross-method comparisons under controlled VLM settings in Table ~\ref{tab:ablation_method}. When utilizing GPT-4, our SeqVLM achieves 44.0 Acc@0.5, outperforming VLM-Grounder’s 32.8 by 11.2 points, confirming fundamental architectural advantages in multimodal fusion and geometric reasoning. Simultaneously, under the Doubao VLM environment, SeqVLM attains 49.6 Acc@0.5, outperforming ZSVG3D's 29.2 and SeeGround's 40.0 by 20.4 and 9.6 points, respectively. These consistent superiority patterns across heterogeneous VLM backends demonstrate that our framework’s innovations in proposal selection, proposal-guided multi-view projection, and iterative reasoning constitute the primary driver of performance gains, rather than mere VLM capability differences. The results establish SeqVLM as a generalized architectural advancement for zero-shot 3D grounding, transferable across modern VLMs.

\paragraph{Hyper-parameter Analysis}
Our ablation reveals that both the VLM capacity threshold $L$ and multi-view frame number $n_{\mathit{frame}}$ exhibit inverse U-shaped effects on accuracy, peaking at $L=4$ and $n_{\mathit{frame}}=5$ (Figures \ref{fig:ablation_L}-\ref{fig:ablation_nframe}). Smaller capacity threshold  ($L=2$) limit cross-candidate contrast through insufficient batch diversity, while larger capacity threshold  ($L\geq6$) overload the VLM's multimodal reasoning. Similarly for, fewer views ($n_{\mathit{frame}}\leq4$) restrict spatial disambiguation and feature richness, whereas excessive views ($n_{\mathit{frame}}\geq6$) introduce redundant or occluded projections as cross-modal noise. The chosen setting ($L=4$, $n_{\mathit{frame}}=5$) provides the best trade-off between information completeness and computational load, maximizing alignment within VLM capacity.

\section{Conclusion}
In this paper, we propose SeqVLM, a novel framework for zero-shot 3D visual grounding that leverages multi-view real-world scene images. Our approach first employs a Proposal Selection Module to eliminate semantically irrelevant proposals, improving both reasoning accuracy and efficiency. Then, we introduce a proposal-guided multi-view projection strategy to mitigate projection misalignment and enrich contextual information across different viewpoints. To enhance the reasoning capabilities of the VLM, we further design an iterative reasoning mechanism. Evaluations on ScanRefer and Nr3D benchmarks demonstrate that SeqVLM achieves state-of-the-art performance in zero-shot 3D visual grounding, matching the accuracy of supervised approaches while requiring no task-specific training.

\section*{Acknowledgments}
This work was supported in part by the National Natural Science Foundation of China under Grant 62176224, Grant 62176092, Grant 62222602, and Grant 62306165; in part by Science and Technology on Sonar Laboratory under grant 2024-JCJQ-LB-32/07 and in part by China Academy of Railway Sciences under Grant 2023Y1357.

\bibliography{samples/reference}

\end{document}

%% file: samples/algos/vlm_predict.tex
\begin{algorithm}[htbp]
    \caption{Interative Reasoning Mechanism}
    \label{algo:vlm-predict} 
    \begin{algorithmic}[1]
        \Require 
            Image sequences $\mathcal{S} = \{ {S}_1, {S}_2, \dots, {S}_n \}$; \
            Query text $T$; \
            Batch size threshold $L$ for VLM.
        \Ensure Index of the target object
        \Function{Predict}{$\mathcal{S}$, $T$, $L$}
            \State $\mathcal{Q} \gets \mathcal{S}$
            \While{$\lvert \mathcal{Q}\rvert > 1$}
                \State $\mathcal{B} \gets$ \Call{Slice}{$\mathcal{Q}$, $L$} \Comment{Slice $\mathcal{Q}$ to batches}
                \State $\mathcal{Q} \gets []$  \Comment{Reset for next round}
                \For{\textbf{each} $B_i$ \textbf{in} $\mathcal{B}$}
                    \State $\mathit{prompt} \gets$ \Call{ConstructPrompt}{$T$, $B_i$}
                    \State $\mathit{index} \gets$ \Call{VLMSelect}{$\mathit{prompt}$}
                    \If{$\mathit{index} \ne \mathrm{None}$}
                        \State $\mathcal{Q} \gets \mathcal{Q} \cup \ \mathcal{S}
                        [\mathit{index}]$
                    \EndIf
                \EndFor
            \EndWhile
            \If{$\mathcal{Q} == \emptyset$}
                \State \Return None
            \EndIf
            \State \Return $\mathcal{Q}\textbf{[0]}.$\textbf{\textit{index}} 
        \EndFunction
    \end{algorithmic}
\end{algorithm}

%% file: samples/tables/scanrefer.tex
\begin{table*}[htbp]
    \caption{Comparative results on ScanRefer. Evaluates 3D visual grounding by scene type: Unique (target is the sole instance of its class) and Multiple (contains same-class distractors).}
    \centering
    \renewcommand{\arraystretch}{1.2} 
    \resizebox{1.0\linewidth}{!}{
        \begin{tabular}{|c|c|c|cc|cc|cc|}
            \hline
            \multirow{2}{*}{\textbf{Method}} & 
            \multirow{2}{*}{\textbf{Source}} & 
            \multirow{2}{*}{\textbf{Modality}} & 
            \multicolumn{2}{c|}{\textbf{Unique}} & 
            \multicolumn{2}{c|}{\textbf{Multiple}} & 
            \multicolumn{2}{c|}{\textbf{Overall}} \\ 
            \cline{4-9}
             &  &  & 
            \textbf{Acc@0.25} & \textbf{Acc@0.5} & 
            \textbf{Acc@0.25} & \textbf{Acc@0.5} & 
            \textbf{Acc@0.25} & \textbf{Acc@0.5} \\ 
            \hline
            \multicolumn{9}{|c|}{\textbf{Fully-Supervised Methods}} \\ 
            \hline
            ScanRefer\cite{scanrefer} & ECCV20 & 3D & 67.64 & 46.19 & 32.06 & 21.26 & 38.97 & 26.10 \\
            TGNN\cite{tgnn} & AAAI21 & 3D & 68.61 & 56.80 & 29.84 & 23.18 & 37.37 & 29.70 \\
            SAT\cite{sat} & ICCV21 & 3D+2D & 73.21 & 50.83 & 37.64 & 25.16 & 44.54 & 30.14 \\
            MVT\cite{mvt} & CVPR22 & 3D+2D & 77.67 & 66.45 & 31.92 & 25.26 & 40.80 & 33.26 \\
            3D-SPS\cite{3dsps} & CVPR22 & 3D+2D & 84.12 & 66.72 & 40.32 & 29.82 & 48.82 & 36.98 \\
            EDA\cite{eda} & CVPR23 & 3D & 85.76 & 68.57 & 49.13 & 37.64 & 54.59 & 42.26 \\
            3DVLP\cite{3dvlp} & CVPR23 & 3D+2D & 84.23 & 64.61 & 43.51 & 33.41 & 51.41 & 39.46 \\
            MCLN\cite{mcln} & ECCV24 & 3D & 86.89 & 72.73 & 51.96 & 40.76 & 57.17 & 45.53 \\ 
            \hline
            \multicolumn{9}{|c|}{\textbf{Zero-Shot Methods}} \\ 
            \hline
            LERF~\cite{kerr2023lerf} & ICCV23 & 3D+2D & - & - & - & - & 4.8 & 0.9 \\
            OpenScene~\cite{openscene} & CVPR23 & 3D+2D & 20.1 & 13.1 & 11.1 & 4.4 & 13.2 & 6.5 \\
            LLM-Grounder~\cite{llm-grounder} & ICRA24 & 3D & - & - & - & - & 17.1 & 5.3 \\
            ZS3DVG~\cite{zs3dvg} & CVPR24 & 3D+2D & 63.8 & 58.4 & 27.7 & 24.6 & 36.4 & 32.7 \\
            VLM-Grounder\cite{vlmgrounder} & CoRL24 & 2D & 66.0 & 29.8 & \textbf{48.3} & 33.5 & 51.6 & 32.8 \\
            SeeGround\cite{li2024seeground} & CVPR25 & 3D+2D & 75.7 & 68.9 & 34.0 & 30.0 & 44.1 & 39.4 \\
            SeqVLM & - & 3D+2D & \textbf{77.3(+1.6)} & \textbf{72.7(+3.8)} & 47.8 & \textbf{41.3(+7.8)} & \textbf{55.6(+4.0)} & \textbf{49.6(+10.2)} \\ 
            \hline
        \end{tabular}
    }
    \label{tab:scanrefer_result}
\end{table*}

%% file: samples/tables/nr3d.tex
\begin{table}[htbp]
    \centering
    \renewcommand{\arraystretch}{1.2} 
    \caption{Comparative results on Nr3D. Evaluates via Easy (single distractor) vs. Hard (multiple distractors) queries and View-Dependent (requires specific viewpoints) vs. View-Independent scenarios.}
    \resizebox{1.0\linewidth}{!}{
        \begin{tabular}{|l|cc|cc|c|}
            \hline
            \textbf{Method} & \textbf{Easy} & \textbf{Hard} & \textbf{Dep.} & \textbf{Indep.} & \textbf{Overall} \\ 
            \hline
            \multicolumn{6}{|c|}{\textbf{Fully-Supervised Methods}} \\ 
            \hline
            ReferIt3D\cite{referit3d} & 43.6 & 37.9 & 32.5 & 37.1 & 35.6  \\
            TGNN\cite{tgnn} & 44.2 & 30.6 & 35.8 & 38.0 & 37.3 \\
            InstanceRefer\cite{instancerefer} & 46.0 & 31.8 & 34.5 & 41.9 & 38.8  \\
            3DVG-Trans\cite{3dvg-trans} & 48.5 & 34.8 & 34.8 & 43.7 & 40.8  \\
            SAT\cite{sat} & 56.3 & 42.4 & 46.9 & 50.4 & 49.2  \\
            EDA\cite{eda} & 58.2 & 46.1 & 50.2 & 53.1 & 52.1 \\
            MCLN\cite{mcln} & - & - & - & - & 59.8  \\ 
            \hline
            \multicolumn{6}{|c|}{\textbf{Zero-Shot Methods}} \\ 
            \hline
            ZS3DVG\cite{zs3dvg} & 46.5 & 31.7 & 36.8 & 40.0 & 39.0 \\ 
            SeeGround\cite{li2024seeground} & 54.5 & 38.3 & 42.3 & 48.2 & 46.1 \\ 
            VLM-Grounder\cite{vlmgrounder} & 55.2 & 39.5 & 45.8 & 49.4 & 48.0 \\ 
            SeqVLM & \textbf{58.1} & \textbf{47.4} & \textbf{51.0} & \textbf{54.5} & \textbf{53.2} \\
            \hline
        \end{tabular}
    }
    \label{tab:nr3d_result}
\end{table}

%% file: samples/tables/abla_module.tex
\begin{table}[htbp]
    \centering
    \begin{threeparttable}
        \caption{Ablation study on different components in our framework on ScanRefer.}
        \label{tab:ablation_module}
        \begin{tabular}{c|cccc|c}
            \toprule
            \textbf{Config} & 
            \textbf{Baseline} & 
            \textbf{PSM\tnote{1}} & 
            \textbf{MVP\tnote{2}} & 
            \textbf{IRM\tnote{3}} & 
            \textbf{Acc@0.5} \\
            \midrule
            1     & \ding{51} &       &       &       & 2.4 \\
            2     & \ding{51} & \ding{51} &       &       & 43.2 \\
            3     & \ding{51} &       & \ding{51} &       & 2.0 \\
            4     & \ding{51} &       &       & \ding{51} & 10.0 \\
            5     & \ding{51} & \ding{51} & \ding{51} &       & 44.4 \\
            6     & \ding{51} &       & \ding{51} & \ding{51} & 10.4 \\
            7     & \ding{51} & \ding{51} &       & \ding{51} & 46.0 \\
            \midrule
            8     & \ding{51} & \ding{51} & \ding{51} & \ding{51} & \textbf{49.6} \\
            \bottomrule
        \end{tabular}
        
        \begin{tablenotes}
            \item[1] PSM: Proposal Selection Module.
            \item[2] MVP: Multi-View Projection.
            \item[3] IRM: Iterative Reasoning Mechanism.
        \end{tablenotes}
    \end{threeparttable}
\end{table}

%% file: samples/tables/abla_vlm.tex
\begin{table}[htbp]
    \centering
    \caption{VLM Performance and Cost .}
        \begin{tabular}{cccc}
            \toprule
            \textbf{VLM} & \textbf{Tokens} & \textbf{Cost} & \textbf{Acc@0.5} \\
            \midrule
            GPT-4 & 1,684k & 2.2\$ & 44.0 \\
            Qwen-vl-max & 1,886k & 0.8\$ & 46.0 \\
            Doubao-1.5-vision-pro & 7,097k & 2.9\$ & \textbf{49.6} \\
            \bottomrule
        \end{tabular}
    \label{tab:ablation_vlm}
\end{table}

%% file: samples/tables/abla_cross_method.tex
\begin{table}[htbp]
    \centering
    \caption{Cross-Method performance comparison under controlled VLM settings.}
    \begin{tabular}{ccc}
        \toprule
        \textbf{Method} & \textbf{LLM/VLM} & \textbf{Acc@0.5} \\
        \midrule
        VLM-Grounder & GPT-4 & 32.8 \\
        SeqVLM & GPT-4 & \textbf{44.0} \\
        ZSVG  & Doubao-1.5-vision-pro & 29.2 \\
        SeeGround & Doubao-1.5-vision-pro & 40.0 \\
        SeqVLM & Doubao-1.5-vision-pro & \textbf{49.6} \\
        \bottomrule
    \end{tabular}
    \label{tab:ablation_method}
\end{table}